\documentclass{article}
\usepackage{spconf,amsmath,graphicx,hyperref,amsfonts}
\usepackage{algorithmic}
\usepackage{graphicx}
\usepackage{textcomp}
\usepackage{xcolor}
\usepackage{longtable}
\usepackage{hyperref}
\usepackage{booktabs}
\usepackage{comment}
\usepackage{multirow}
\usepackage{tabularx}

\title{CE-GOCD: Central Entity-Guided Graph Optimization for Community Detection to Augment LLM Scientific Question Answering}
\name{Jiayin Lan$^{\star}$, Jiaqi Li$^{\dagger}$$^{\ddagger}$, Baoxin Wang$^{\dagger}$, Ming Liu$^{\star}$$^{\ddagger}$, Dayong Wu$^{\dagger}$, Shijin Wang$^{\dagger}$, Bing Qin$^{\star}$,Guoping Hu$^{\dagger}$}
\address{$^{\star}$Harbin Institute of Technology, Harbin, China\\
$^{\dagger}$State Key Laboratory of Cognitive Intelligence, iFLYTEK Research, China\\
jqli@tjnu.edu.cn, mliu@ir.hit.edu.cn, $^{\ddagger}$Corresponding author}
\begin{document}
%
\maketitle
\begin{abstract}
Large Language Models (LLMs) are increasingly used for question answering over scientific research papers. Existing retrieval augmentation methods often rely on isolated text chunks or concepts, but overlook deeper semantic connections between papers. This impairs the LLM's comprehension of scientific literature, hindering the comprehensiveness and specificity of its responses. To address this, we propose Central Entity-Guided Graph Optimization for Community Detection (CE-GOCD), a method that augments LLMs' scientific question answering by explicitly modeling and leveraging semantic substructures within academic knowledge graphs. Our approach operates by: (1) leveraging paper titles as central entities for targeted subgraph retrieval, (2) enhancing implicit semantic discovery via subgraph pruning and completion, and (3) applying community detection to distill coherent paper groups with shared themes. We evaluated the proposed method on three NLP literature-based question-answering datasets, and the results demonstrate its superiority over other retrieval-augmented baseline approaches, confirming the effectiveness of our framework.
\end{abstract}
\section{Introduction}
\label{sec:intro}

\begin{figure}[htb]
  \small
  \centering
  \includegraphics[width=0.85\columnwidth]{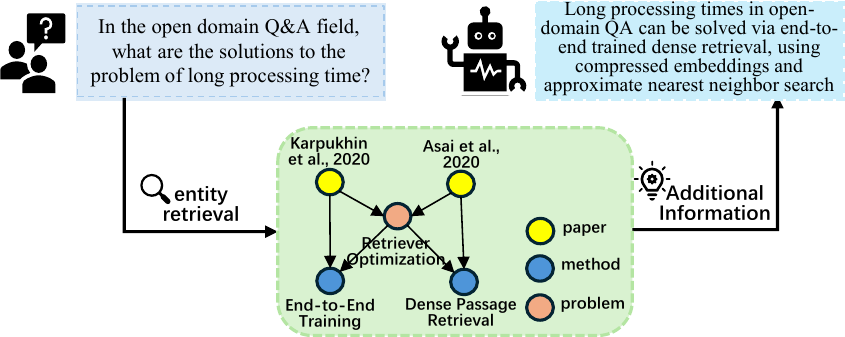}
\caption{Knowledge graph augmentation for scientific QA.}
\label{fig:construction}
\end{figure}

In the field of scientific literature, Large Language Models (LLMs) like GPT4 have been widely used for tasks such as literature retrieval \cite{ajith-etal-2024-litsearch}, review generation \cite{agarwal2024litllm}, and literature analysis \cite{wu-etal-2024-sparkra}. To mitigate issues including lack of flexibility \cite{razdaibiedina2023progressive}, hallucination \cite{ji2023survey}, and professionalism \cite{ling2023domain} in scientific question answering, researchers often augment LLMs with external knowledge like relevant texts or structured knowledge graphs \cite{zhao-etal-2024-enhancing}  (as shown in Figure \ref{fig:construction}).

Existing retrieval augmentation methods struggle to address scientific questions that require integration of both paper-specific elements and conceptual knowledge. Document retrieval techniques \cite{lewis2020retrieval} are challenging for multi-source information fusion, and often overlook deeper inter-document relations. Knowledge graph based approaches \cite{mondal-etal-2021-end} excel at capturing conceptual relations but lack a deep exploration of the implicit connections both between concept entities and within scientific literature. These shortcomings can lead to incomplete or inaccurate answers.

Leveraging the structure of knowledge graphs to identify semantic connections among academic papers has become an important topic. The goal is to enable LLMs to grasp the core ideas of relevant publications more comprehensively and systematically, thereby producing coherent responses. To address this challenge, we propose Central Entity-Guided Graph Optimization for Community Detection method operating on an NLP article knowledge graph.

Since each paper forms an independent semantic unit, we take the title as its central entity, connected to all paper components. Relevance subgraphs are built by extracting paths between question-related entities and title-linked nodes, capturing inter-paper relational structures. These subgraphs explicitly represent research entities and paper elements, providing structured semantic context. We refine them by pruning redundant paths and inferring implicit correlations between entities. Finally, title-guided community detection identifies clusters of semantically coherent publications, enabling structured knowledge aggregation for relational reasoning. 

\begin{figure*}[htb]
    \centering
    \includegraphics[width=0.8\textwidth]{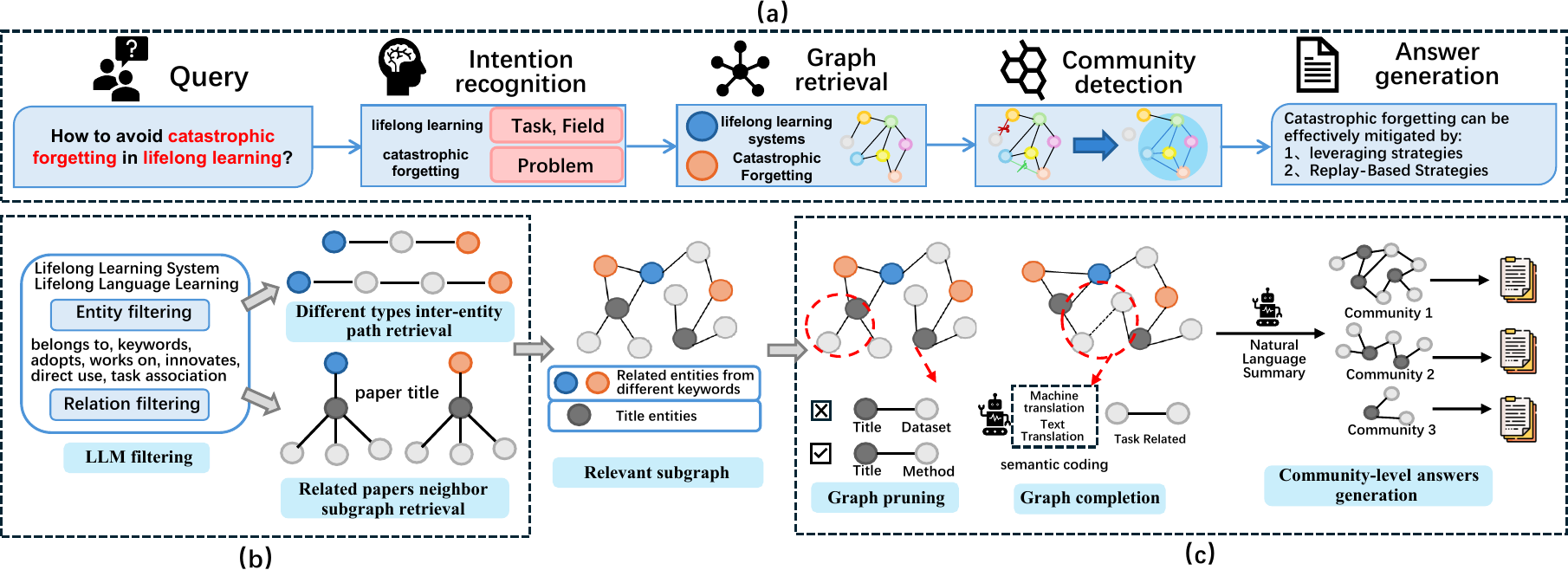}
    \caption{(a) CE-GOCD chematic diagram. We perform graph retrieval (b) using LLM-based filtering along with inter-entity path and neighbor subgraph extraction. This is followed by subgraph optimization and community detection (c), which involve graph pruning, graph completion, and community-level answers generation.}
    \label{fig:method}
\end{figure*}

We evaluate CE-GOCD's question-answering performance against state-of-the-art retrieval augmentation baselines and LLMs across three benchmark datasets: QASPER \cite{dasigi-etal-2021-dataset}, PeerQA \cite{baumgartner-etal-2025-peerqa}, and  NLP-AKG, a multi-paper question answering dataset we annotated in NLP domain. The experimental results demonstrate that our method achieves superior performance compared to baseline approaches.

\section{PROPOSED METHOD CE-GOCD}
\label{sec:pagestyle}
Our approach CE-GOCD innovates by integrating title-guided subgraph retrieval, pruning and completion for implicit relation discovery, and title-based community detection to structurally aggregate knowledge from academic publications. The chematic diagram is shown in Figure \ref{fig:method}.
\subsection{Subgraph Retrieval}
Our method employs LLM to extract from the query $Q$ a set of contextual keywords $K = \{k_1, k_2, \dots, k_m\}$ that serve as contextual conditions and target entity types $T = \{t_1, t_2, \dots, t_n\}$ that are potentially involved in the answer.

For each keyword $k_i \in K$, we retrieve the top-10 related entities $E_i$ from the knowledge graph via TF-IDF matching. Based on the entity types, we identify a set of candidate relation types $R$. LLM then filters these entities and relations to remove irrelevant elements, yielding refined sets $\widetilde{E}$ and $\widetilde{R}$.

Relevant entities originating from different keywords are paired: $P = \{(e_a, e_b) \mid e_a \in E_i, e_b \in E_j, i \ne j\}$. For each pair $(e_a, e_b) \in P$, paths are retrieved with the filtered relations $r_i \in \widetilde{R}$. Our hyperparameter experiments determined 5 hops as the optimal limit of paths, beyond which recall drops due to noise introduction and response time rises excessively:
\begin{equation}
\text{Path}(e_a, e_b) = (e_a, r_1, e_1, \dots, r_k, e_b),\quad k \leq 5
\end{equation}

Since paper title nodes $E_{\text{title}}$ index all relevant paper elements, we include $E_{\text{title}}$ corresponding to the relevant entities $\widetilde{E}$, along with their direct connections to target type entities $E_T$, in the retrieved neighbor subgraph. The final retrieved relevant subgraph contains all entities and relations from the relevant paired entity paths and the neighbor subgraphs.

\subsection{Subgraph Optimization }
\label{sec:print}
\begin{table*}[h]
    \small 
    \centering
    \caption{Performance of our method and baselines on three datasets.}
    \begin{tabular}{l|l|ccc|ccc|ccc}
\toprule
\textbf{Method} & \textbf{LLM} & \multicolumn{3}{c|}{\textbf{NLP-MQA}} & \multicolumn{3}{c|}{\textbf{peerQA}} & \multicolumn{3}{c}{\textbf{QASPER}} \\
\cmidrule(lr){3-5} \cmidrule(lr){6-8} \cmidrule(l){9-11}
& & \textbf{Pre} & \textbf{Rec} & \textbf{F1} & \textbf{Pre} & \textbf{Rec} & \textbf{F1} & \textbf{Pre} & \textbf{Rec} & \textbf{F1} \\
\midrule
BM25 & GPT-4 & 0.6890 & 0.7097 & 0.6989 & 0.6794 & 0.7386 & 0.7048 & 0.7462 & 0.7550 & 0.7487 \\
Embedding retrieval & GPT-4 & 0.6878 & 0.7117 & 0.6992 & 0.6829 & 0.7383 & 0.7069 & 0.7508 & 0.7607 & 0.7542 \\
KAG & GPT-4 & 0.6819 & 0.7797 & 0.7275 & 0.6873 & 0.7387 & 0.7097 & 0.6629 & 0.7703 & 0.7117 \\
PathRAG & GPT-4 & 0.7220 & 0.7789 & 0.7482 & 0.5916 & 0.7276 & 0.6511 & 0.6398 & 0.7423 & 0.6850 \\
MindMap & GPT-4 & 0.7628 & 0.7386 & 0.7496 & 0.6883 & 0.7335 & 0.7080 & 0.7430 & 0.7359 & 0.7380 \\
\hline
LLM only & GPT-4 & 0.6985 & 0.7039 & 0.7006 & 0.6709 & 0.7220 & 0.6929 & 0.7631 & 0.7464 & 0.7531 \\
CE-GOCD (ours) & GPT-4 & \textbf{0.7666} & \textbf{0.8156} & \textbf{0.7901} & \textbf{0.7546} & \underline{\textbf{0.7683}} & \underline{\textbf{0.7605}} & \textbf{0.7744} & \textbf{0.8038} & \textbf{0.7869} \\
\hline
LLM only & DeepSeek-V3 & 0.7091 & 0.7534 & 0.7303 & 0.6932 & \textbf{0.7578} & 0.7200 & 0.6277 & 0.7706 & 0.6908 \\
CE-GOCD (ours) & DeepSeek-V3 & \underline{\textbf{0.7830}} & \textbf{0.8211} & \underline{\textbf{0.8014}} & \textbf{0.7487} & 0.7408 & \textbf{0.7435} & \textbf{0.7810} & \textbf{0.7984} & \textbf{0.7884} \\
\hline
LLM only & Qwen-plus & 0.7231 & 0.7461 & 0.7338 & 0.6892 & 0.6952 & 0.6903 & 0.7559 & 0.7553 & 0.7539 \\
CE-GOCD (ours) & Qwen-plus & \textbf{0.7720} & \underline{\textbf{0.8246}} & \textbf{0.7973} & \underline{\textbf{0.7552}} & \textbf{0.7467} & \textbf{0.7495} & \underline{\textbf{0.8293}} & \underline{\textbf{0.8342}} & \underline{\textbf{0.8309}} \\
\bottomrule
\end{tabular}
    \label{tab:table1}
\end{table*}
This part consists of two steps: \textbf{pruning} and \textbf{completion}. 

\textbf{Pruning} removes edges with weak relevance to query $Q$ and their surrounding entities, while also assigning edge weights for subsequent community detection.

The weight $w_{ij}$ of an edge connecting $e_i$ and $e_j$ is determined by:
\begin{itemize}
    \item the semantic similarity $S_{\text{semantic}}$ between the triplet $(e_i, r, e_j)$ and the keyword set $K$;
    \item the type-based weight $W_{\text{type}}(r)$ of relation $r$, assigned with LLM assistance.
\end{itemize}

The combined edge weight is:
\begin{equation}
w_{ij} = S_{\text{semantic}}((e_i, r, e_j), K) \times W_{\text{type}}(r)
\end{equation}

A self-adaptive threshold $\theta$ is employed, which is adjusted according to the size of the subgraph. Edges with $w_{ij} < \theta$ and isolated nodes are pruned, yielding a refined subgraph $\widetilde{G}$.

\textbf{Completion} is to reveal implicit relations (e.g., similar tasks or methodologies) within the subgraph. We process entities of each type $t \in T$ via semantic encoding and proximity analysis. For type $t$, let $E_t = \{e_1, e_2, \dots, e_n\}$ be its entities. Each entity $e_i$ is encoded into an embedding $v_i \in \mathbb{R}^d$ using a pre-trained language model. These embeddings are projected into a 1D semantic space, yielding coordinates $z_i \in \mathbb{R}$. The Euclidean distance between these coordinates reflects the semantic relevance between the corresponding entities.

A distance threshold $\theta$ is determined from the differences:
\begin{equation}
\quad \theta = Q_{0.75}(\Delta) + 1.0 \cdot \text{IQR}(\Delta),
\Delta = \{\delta_i = |z_{i+1} - z_i|\}
\end{equation}

 Here, IQR is the interquartile range, a robust measure of data spread. Candidate pairs with strong semantic relevance are identified with the threshold $\theta$:

\begin{equation}
\mathcal{P} = \left\{ (e_i, e_j) \middle| |z_i - z_j| \leq \theta,  i \neq j \right\}
\end{equation}

The LLM identifies hidden correlations among the candidate pairs, and relevance weights $w_{ij}$ are computed for the new relations before they are integrated into the subgraph $\widetilde{G}$.

\subsection{Community Detection}

The relevant subgraph contains rich connections among papers and elements. To facilitate comprehension, we cluster papers into communities with high internal relevance, reflecting shared aspects related to query $Q$.

We first partition the graph into cohesive communities using modularity-based clustering. Let $w_{ij}$ be the weight of edge between entity nodes $e_i$ and $e_j$. The Louvain \cite{Blondel_2008} algorithm maximizes modularity:
\begin{equation}
Q = \frac{1}{2m}\sum_{ij}\left[w_{ij} - \frac{k_i k_j}{2m}\right]\delta(c_i, c_j)
\end{equation}
where $m = \sum_{ij} w_{ij}$, $k_i = \sum_j w_{ij}$, and $\delta(c_i, c_j) = 1$ if $e_i$ and $e_j$ belong to the same community, else 0.

The initial communities $\mathcal{C} = \{C_1, C_2, \dots, C_k\}$ are refined through iterative merging until $|\mathcal{C}| \leq \theta_{\text{max}}$, where $\theta_{\text{max}}$ is the maximum allowed communities (typically 3).

To ensure that the division of the community is dominated by the paper, for each community $C_i$, we identify a central title node $e_\text{title}^i$ with the highest weighted degree, where $E_\text{title}^i \subseteq C_i$ represents paper title nodes within community $C_i$:
\begin{equation}
e_\text{title}^i = \arg\max_{e \in E_\text{title}^i} \sum_{v \in C_i} w(t, v)
\end{equation}

Relations between entities are translated into natural language descriptions, enabling the LLM to refine community division and summarize shared themes. These are then provided to the LLM to generate community-level answers by integrating intra-community information, which are finally synthesized into a comprehensive final answer.

\section{EXPERIMENT}
\subsection{Relevant Dataset}
This work is based on NLP-AKG \cite{lan2025nlpakgfewshotconstructionnlp}, a large-scale academic knowledge graph for natural language processing. Constructed from 61,826 ACL Anthology papers (1952–2024), it contains 620,353 entities and 2,271,584 automatically extracted relations across 15 types (e.g., models, datasets) and 29 semantic and citation relation types. NLP-AKG integrates concepts, title-anchored publications, metadata, and citations to uncover research trends and scholarly influence in NLP.

The evaluation is conducted on three datasets: QASPER \cite{dasigi-etal-2021-dataset}, a benchmark for long-document question answering; PeerQA \cite{baumgartner-etal-2025-peerqa}, which collects questions from academic peer reviews with answers provided by the original paper authors; and our annotated NLP-MQA, a 200 NLP domain question-answer pairs dataset, due to the lack of open-source datasets in NLP multi-paper question answering. 
\subsection{Implementation Details}
We conduct comprehensive comparisons with multiple baselines, including knowledge graph augmentation methods (KAG\cite{liang2024kag}, MindMap \cite{wen-etal-2024-mindmap}, PathRAG \cite{chen2025pathragpruninggraphbasedretrieval}), and two retrieval-based methods (BM25 and text embedding retrieval). We use BERTScore \cite{zhang2020bertscoreevaluatingtextgeneration} to measure the semantic similarity between the generated and the reference answer.

\subsection{Experimental Results}
Based on the comprehensive experimental results, our proposed CE-GOCD method demonstrates improvements across different datasets and LLMs, highlighting its robustness and effectiveness. Table \ref{tab:table1} presents the results, with the best performance per dataset underlined.

On our manually annotated NLP-MQA, CE-GOCD achieved the highest F1-score across all LLMs: 79.01\% with GPT-4 (8.95\% improvement over LLM-only), 80.14\% with DeepSeek-V3 (+7.11\%), and 79.73\% with Qwen-Plus (+6.35\%). On PeerQA, CE-GOCD with GPT-4 yielded an F1-score improvement of +6.76\% over LLM-only, and with Qwen-Plus showed a gain of +5.92\%. On QASPER, the F1 improvement reached +9.76\% with DeepSeek-V3 and +7.70\% with Qwen-Plus, demonstrating the robustness of our approach across different models and datasets.

Compared to traditional retrieval and KG-enhanced methods, CE-GOCD maintains high precision while significantly improving recall, validating the strong generalizability of our knowledge graph strategy across varied datasets and LLMs for cross-document question answering. 

We also evaluated the computational cost, with results showing an average response time of 129 seconds, processing approximately 7,208 input tokens and generating 3,759 output tokens, balancing depth of analysis with resource use.

\subsection{Domain Generalizability Experimental Results}
To assess the generalizability of our method in knowledge graph-based LLM question answering, we employed the ChatDoctor5K \cite{li2023chatdoctormedicalchatmodel} medical dataset to construct a domain-specific knowledge graph containing diseases, symptoms, and diagnoses. Disease nodes served as central entities for subgraph extraction. The results are
shown in Table \ref{tab:table2}.
\begin{table}[htbp]
    \small
    \centering
    \caption{Generalizability experimental results of our method.}
    \begin{tabular}{lccc}
\toprule
\textbf{Model} & \textbf{Pre} & \textbf{Rec} & \textbf{F1} \\
\midrule
GPT-4 & 0.7689 & 0.7893 & 0.7786 \\
MindMap & 0.7936 & 0.7977 & 0.7954 \\
CE-GOCD (ours)  & 0.7674 & 0.8167 & 0.7910 \\
\bottomrule
\end{tabular}
    \label{tab:table2}
\end{table}

Although our method does not outperform all baselines due to differences in conceptual structures between scientific and medical domains, its narrow gap with MindMap demonstrates strong potential for generalization across fields.
\subsection{Ablation Experimental Results}

To evaluate the design validity of each model component, we perform ablation studies on the annotated NLP-MQA dataset. The proposed approach is assessed under the following two conditions. The results are shown in Table \ref{tab:table3}.
\begin{table}[htbp]
    \small
    \centering
    \caption{Ablation experimental results of our method.}
    \begin{tabular}{lccc}
\toprule
\textbf{Model Variant} & \textbf{Pre} & \textbf{Rec} & \textbf{F1} \\
\midrule
CE-GOCD -SO & 0.5653 & 0.6618 & 0.6093 \\
CE-GOCD -CD & 0.7243 & 0.8080 & 0.7637 \\
CE-GOCD & \textbf{0.7830} & \textbf{0.8211} & \textbf{0.8014} \\ 
\bottomrule
\end{tabular}
    \label{tab:table3}
\end{table}

 To evaluate the role of subgraph optimization, we conducted community mining and answer generation directly on the unrefined subgraph (-SO). For community detection, we supplied the LLM with the processed graph without clustering (-CD). The results confirmed that structured refinement is crucial for capturing relational semantics among concepts, and grouping papers into thematic communities considerably improves literature identification and response organization.

\subsection{Graph Optimization Bias Analysis}
\begin{figure}[htb]
    \small
    \centering
    \includegraphics[width=0.9\columnwidth]{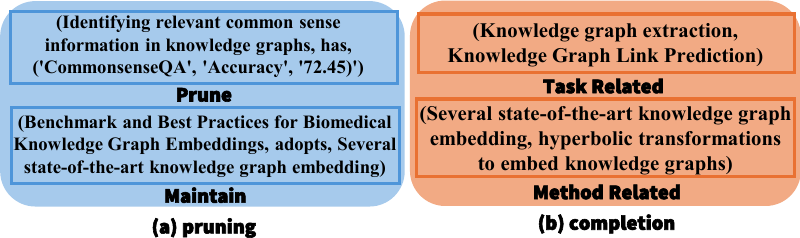}
    \caption{An example of pruning (a) and completion (b).}
    \label{fig:example}
\end{figure}
We analyzed triplets before and after pruning and completion to confirm no errors or biases were introduced in the pruning and completion processes (as shown in Figure \ref{fig:example}). We compared triplet quantity, accuracy, and relevance. Accuracy (correct triplets proportion) and relevance (related triplets proportion) were evaluated separately by both LLM and human expert,  then averaged with a Kappa value of at least 0.8. The results are presented in Table \ref{tab:table4}.

\begin{table}[htbp]
    \small
    \centering
    \caption{Triple metrics before and after graph processing.}
    \begin{tabular}{lccc}
\toprule
\textbf{State} & \textbf{Quantity} & \textbf{Accuracy} & \textbf{Relevance} \\
\midrule
Before processing & 162 & 0.952 & 0.478 \\
After pruning & 94 & 0.943 & 0.668 \\
After completion & 131 & 0.939 & 0.661 \\
\bottomrule
\end{tabular}
    \label{tab:table4}
\end{table}
The pruning and completion processes not only enrich the number of relations in the relevant subgraph but also enhance their relevance while maintaining accuracy. This results in a structurally more coherent and better-connected subgraph, which facilitates more effective community detection.
\section{Conclusions}
This paper proposes CE-GOCD, a novel framework that integrates graph operations and community-driven knowledge augmentation. Supported by a constructed NLP academic knowledge graph, the method effectively captures inter-paper relations and conceptual structures. Experimental results across multiple benchmarks demonstrate that CE-GOCD outperforms current state-of-the-art retrieval augmentation approaches and LLMs, offering more comprehensive and accurate answers in scientific question-answering scenarios.
\section{Acknowledgment}
The research in this article is supported by the New Generation of Artificial Intelligence National Science and Technology Major Project (2023ZD0121503), the National Science Foundation of China (U22B2059, 62276083), Key Research and Development Program of Heilongjiang Providence (2024ZX01A05), and the 5G Application Innovation Joint Research Institute’s Project (A003).
\vfill\pagebreak

\bibliographystyle{IEEEbib}
\bibliography{refs}

\end{document}